%% file: main.tex
\documentclass[letterpaper, 10 pt, conference]{ieeeconf}  

\IEEEoverridecommandlockouts                              

\usepackage{graphics} 
\usepackage{epsfig} 
\usepackage{stfloats}
\usepackage{multirow}
\usepackage{multicol}
\usepackage{colortbl}
\usepackage{amsmath}
\definecolor{myyel}{rgb}{.99,.92,.41}
\definecolor{mycy}{rgb}{.99,0.5,0.6}
\usepackage{amsmath} 
\usepackage{amssymb}  
\usepackage{newtxmath}
\usepackage{cite}
\usepackage{multirow}
\usepackage{bbding}
\usepackage{url}
\usepackage{algorithmic}
\usepackage{algorithm}
\usepackage{hyperref}
\hypersetup{hidelinks,
	colorlinks=true,
	allcolors=black,
	pdfstartview=Fit,
	breaklinks=true}
\title{\LARGE \bf
GSORB-SLAM: Gaussian Splatting SLAM benefits from ORB features and Transmittance information
}

\author{Wancai Zheng$^{1}$, Xinyi Yu$^{1,*}$, Jintao Rong$^{1}$, Linlin Ou$^{1}$, Yan Wei$^{1}$, Libo Zhou$^{1}$
\thanks{This research was supported by the Baima Lake Laboratory Joint Funds of the Zhejiang Provincial Natural Science Foundation of China under Grant No. LBMHD24F030002 and the National Natural Science Foundation of China under Grant 62373329.}
\thanks{$^{1}$All authors are with the School of Information Engineering, Zhejiang University of Technology, Hangzhou, China, 310000.
        {\tt\small \{aczheng,yuxy\}@zjut.edu.cn}}%
\thanks{* Corresponding author}%
}

\begin{document}

\maketitle
\thispagestyle{empty}
\pagestyle{empty}


\begin{abstract}
The emergence of 3D Gaussian Splatting (3DGS) has recently ignited a renewed wave of research in dense visual SLAM. However, existing approaches encounter challenges, including sensitivity to artifacts and noise, suboptimal selection of training viewpoints, and the absence of global optimization. 
In this paper, we propose GSORB-SLAM, a dense SLAM framework that integrates 3DGS with ORB features through a tightly coupled optimization pipeline. To mitigate the effects of noise and artifacts, we propose a novel geometric representation and optimization method for tracking, which significantly enhances localization accuracy and robustness. For high-fidelity mapping, we develop an adaptive Gaussian expansion and regularization method that facilitates compact yet expressive scene modeling while suppressing redundant primitives. 
Furthermore, we design a hybrid graph-based viewpoint selection mechanism that effectively reduces overfitting and accelerates convergence. Extensive evaluations across various datasets demonstrate that our system achieves state-of-the-art performance in both tracking precision—improving RMSE by 16.2\% compared to ORB-SLAM2 baselines—and reconstruction quality—improving PSNR by 3.93 dB compared to 3DGS-SLAM baselines. The project: \url{https://aczheng-cai.github.io/gsorb-slam.github.io/}
\end{abstract}

\input{section/1_introduction}

\input{section/2_relatedwork}

\input{section/3_method}

\input{section/4_experiment}

\input{section/5_conclusion}



\end{document}

%% file: section/1_introduction.tex
\section{INTRODUCTION}
Over the past two decades, Simultaneous Localization and Mapping (SLAM) has remained a prominent research topic, evolving from traditional SLAM, which emphasizes improving localization accuracy, to neural radiance field (NeRF) SLAM, which offers rich scene representations. As a critical component in fields such as autonomous driving, virtual reality (VR), and embodied artificial intelligence~\cite{embodiedai}, SLAM has gained increasing significance in map representation, which is essential for downstream tasks. Visual SLAM has introduced various map representations, including point/surfel clouds~\cite{multiresolution,elasticFusion,orbslam3}, mesh representations~\cite{surfelmeshing,ovpcmesh}, and voxels~\cite{efficient,kinectfusion}.

Recently, NeRF-based novel view synthesis~\cite{nerf} has garnered widespread attention among researchers due to its high-fidelity output, leading to the development of numerous advanced dense neural SLAM methods that have made significant progress. However, the computational expense of volumetric rendering in NeRF limits its ability to produce full-resolution images, resulting in outputs that fall short of the desired photorealism.

Notably, the 3D Gaussian splatting (3DGS) technique~\cite{3dgs} has enabled high-quality, full-pixel novel view synthesis and real-time scene rendering within standard GPU-accelerated frameworks. Consequently, several SLAM methods based on 3DGS~\cite{photoslam,gsslam,bridge} have emerged, significantly enhancing rendering quality and achieving rendering speeds up to 100 times faster than NeRF-based SLAM. Nonetheless, pressing challenges remain, such as the Bundle Adjustment (BA) problem and sensitivity to artifacts, which can degrade tracking accuracy. Additionally, the absence of multi-view constraints and strong anisotropy often leads to Gaussian overfitting to the current viewpoint.

To address these challenges, we present GSORB-SLAM, a tightly coupled 3DGS and ORB feature SLAM system. We develop an adaptive extended Gaussian strategy for dense mapping that dynamically initializes Gaussians in under-reconstructed regions by integrating accumulated transmittance analysis with geometric cues, facilitating the construction of higher-quality maps. Furthermore, to mitigate overfitting resulting from insufficient multi-view constraints, we propose a hybrid graph-based viewpoint selection mechanism that combines overlap graphs with co-visibility graphs, effectively enhancing rendering quality and convergence speed. In the tracking module, we employ a two-stage approach comprising frame-to-frame and frame-to-model modes. The former provides an initial coarse estimate for the latter, which is subsequently refined by integrating joint photometric-reprojection error minimization with surface depth, a novel geometric representation for tracking introduced in this paper, thus improving tracking robustness. Ultimately, the presence of feature points enables us to perform lightweight backend BA, reducing accumulated errors and alleviating the computational burden on the device. Our main contributions are summarized as follows:
\begin{itemize}
\item We propose an adaptive extended differentiable Gaussian strategy along with a rendering frame selection mechanism that utilizes a hybrid graph to achieve high-fidelity and compact scene representations.

\item We propose a method that utilizes accumulated transmittance surface depth in conjunction with the joint optimization of reprojection and photometric errors, thereby further enhancing tracking performance.

\item We conduct experiments on various datasets, and the results indicate that our method surpasses the baseline in both tracking and mapping.
\end{itemize}

\begin{figure*}[t]
    \centering
    \includegraphics[scale=0.6]{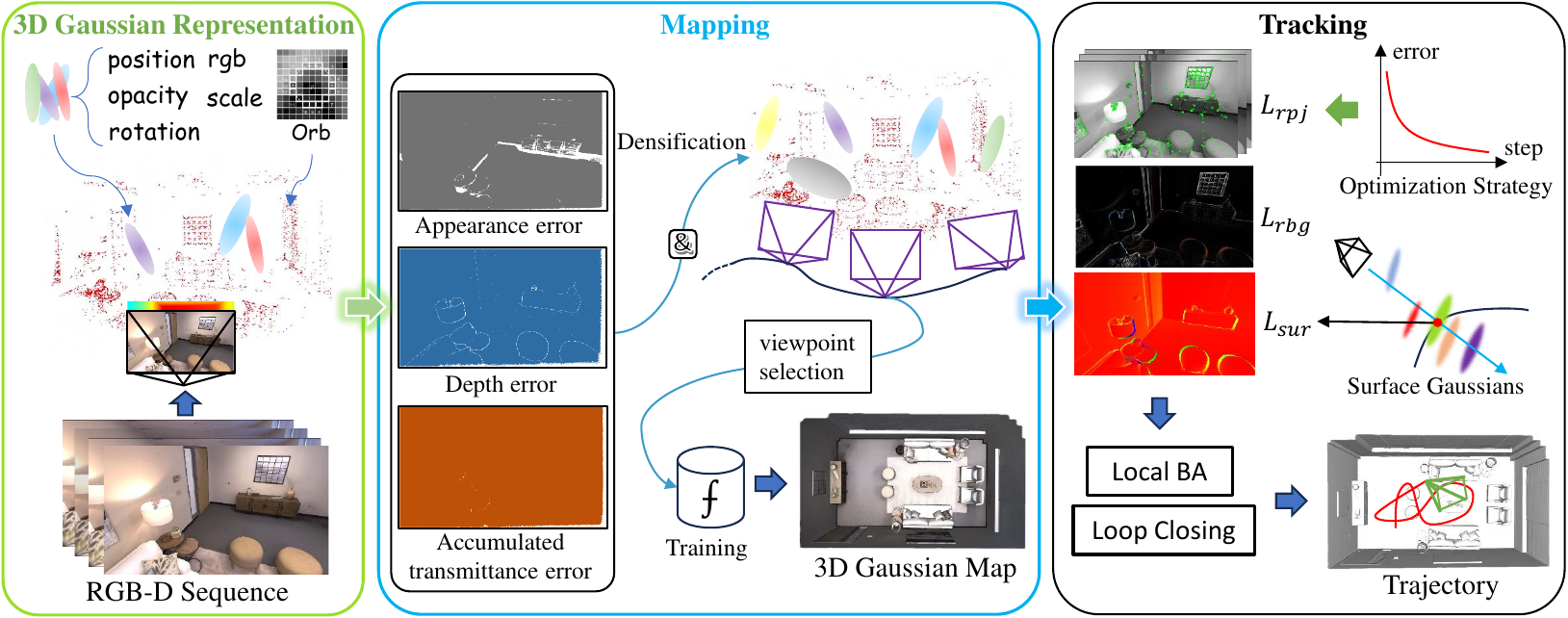}
    \caption{In the 3D Gaussian representation, RGB-D sequences serve as input. Gaussians are generated in the scene based on the re-rendering of color, depth, and transmittance. A viewpoint selection strategy is employed to select the rendering frames for map training. Furthermore, we tightly couple feature points with surface depth and conduct lightweight global optimization using sparse point clouds, ultimately achieving accurate and robust tracking.}
    \vspace{-6mm}
    \label{fig:pipeline}
\end{figure*}

%% file: section/2_relatedwork.tex
\section{RELATED WORK}
\subsection{NeRF SLAM}
Neural Radiance Fields (NeRF)~\cite{nerf} achieves realistic image rendering through pixel-ray sampling of images. By employing a multi-layer perceptron and volumetric differentiable rendering, NeRF facilitates novel view synthesis of unobserved regions. iMAP~\cite{imap} is the first method to integrate NeRF into SLAM, achieving implicit neural representation. Since then, several advanced NeRF-based SLAM methods~\cite{eslam,coslam,uncleslam} have been developed. NICE-SLAM~\cite{niceslam} introduces a hierarchical multi-feature grid to enhance high-quality scene reconstruction, updating only the visible grid features at each step, in contrast to iMAP. Orbeeze-SLAM~\cite{orbeezslam} utilizes ORB features for triangulation, enabling monocular tracking without a depth estimator, while also leveraging NeRF for implicit scene representation. Point-SLAM~\cite{pointslam} distributes point clouds onto object surfaces, demonstrating improved reconstruction quality through neural point cloud representation. However, the ray sampling process in NeRF for individual pixels incurs significant computational costs, resulting in rendering speeds of less than 15 frames per second (FPS).

\subsection{3DGS SLAM}
Recently, 3D Gaussian Splatting (3DGS) has achieved significantly faster rendering speeds, reaching up to 300 FPS, by employing tile-based rasterization for high-resolution image rendering. Many researchers have sought to integrate this advanced technology into SLAM~\cite{gaussiansplattingslam,sgsslam,cgslam,ngmslam}. SplatTAM~\cite{splatam} optimizes camera poses by minimizing photometric error through differentiable Gaussians guided by silhouettes, while adding 3DGS to unobserved areas based on the geometric median depth error. Nevertheless, this approach can lead to an excessive number of Gaussian primitives, significantly increasing memory requirements and computational burden. Gaussian-SLAM~\cite{gaussianslam} explores the limitations of 3DGS in SLAM, utilizing submaps to seed and optimize Gaussians, thereby achieving camera pose tracking and map rendering. GS-SLAM~\cite{gsslam} derives an analytical formula for backward optimization with re-rendering of RGB-D loss and employs depth filtering to eliminate unstable Gaussians during tracking. However, establishing the appropriate depth filtering threshold poses challenges, as it must accommodate noise and convergence levels. Photo-SLAM~\cite{photoslam} uses feature points to determine camera poses and proposes a Gaussian-pyramid-based training approach for scene construction. TANBRIDGE~\cite{bridge} develops a method that connects ORB-SLAM3 with 3DGS, facilitating the integration of these techniques. Although both Photo-SLAM~\cite{photoslam} and TANBRIDGE~\cite{bridge} employ feature points, they do not fully leverage the potential performance benefits of these features. Our analysis of 3DGS indicates that feature points can significantly enhance 3DGS-SLAM in various aspects.

\subsection{Traditional SLAM}
Traditional visual SLAM methods have evolved into two primary categories: direct methods and feature methods. Direct methods~\cite{dsoslam, lsdslam} estimate camera poses by minimizing photometric error and comparing pixel differences between adjacent frames, relying on a strong assumption of photometric consistency. In contrast, feature methods~\cite{monoslam,plslam} utilize image features such as corners and lines, minimizing pixel errors between corresponding features across frames using reprojection error. The ORB-SLAM series~\cite{orbslam,orbslam2,orbslam3} achieves real-time accurate localization based on ORB features and performs global optimization of sparse landmarks and poses through graph optimization. While these methods excel in real-time performance and localization accuracy compared to advanced scene representation techniques (e.g., NeRF, 3DGS), their limited environmental representation capabilities constrain their applicability to more sophisticated tasks.

%% file: section/3_method.tex
\section{METHOD}
The overview of our propose GSORB-SLAM system is illustrated in Fig.~\ref{fig:pipeline}. Given a set of RGB-D sequences as input, we first introduce a 3D Gaussian representation, denoted as $\bf{G}$, and describe the acquisition of surface depth (see \ref{sub:Preliminary}). Next, we present an efficient method for utilizing Gaussians in incremental mapping and outline the criteria for selecting a rendering frame (RF) for training (refer to \ref{sub:Mapping}). Finally, we detail how to achieve accurate and robust pose tracking by jointly employing ORB features and lightweight global optimization based on sparse landmarks (see \ref{sub:Tracking}).

\subsection{3D Gaussian Representation}
\label{sub:Preliminary}
\textbf{Gaussian map representation.} Our scene of SLAM is represented by a collection of anisotropic 3D Gaussian with opacity attributes:
\begin{equation}
{\bf{G}} = \left\{ {{G_i}:({X^w_i},{\Sigma _i},{o_i},{c_i})|i = 1,...,N} \right\}.
\end{equation}
Each Gaussian, as noted in ~\cite{3dgs}, is characterized by a set of parameters: the position ${X^w_i} \in {\mathbb{R}^3}$ in the world coordinate system, the opacity ${o_i} \in [0,1]$, the trichromatic vector ${c_i} \in {\mathbb{R}^3}$ which represents the color of each Gaussian, and the 3D covariance matrix ${\Sigma _i} \in {\mathbb{R}^{3 \times 3}}$: 
\begin{equation}
\Sigma_i  = RS{S^T}{R^T}.
\end{equation}
This covariance matrix is composed of a rotation matrix  $R \in {SO(3)}$ and a scale diagonal matrix $S \in {\mathbb{R}^{3 \times 3}}$.

\textbf{Differentiable rendering.} The 3DGS technique renders colors by blending Gaussian distributions along rays, progressing from the near field to the far field, and subsequently projecting them onto the pixel screen. The color rendering can be mathematically expressed as follows:
\begin{equation}
\tilde C = \sum\limits_{i = 1}^n {{c_i}{\alpha _i}{T_i}}, {T_i} = \prod\limits_{j = 1}^{i - 1} {(1 - {\alpha _j})},
\end{equation}
where the density ${\alpha _i}$ represents the opacity contribution of each Gaussian to the pixel, which is determined by both the Gaussian function and its opacity. The term ${T_i}$ denotes the accumulated transmittance, which decreases as the ray traverses through multiple Gaussians. In order to capture geometric information, the depth $z$ in the camera coordinate system is utilized instead of the color $c$ for rendering:
\begin{equation}
\tilde D = \sum\limits_{i = 1}^n {{z_i}{\alpha _i}{T_i}}. 
\end{equation}

\begin{figure}[t]
\centering

\includegraphics[scale=0.9]{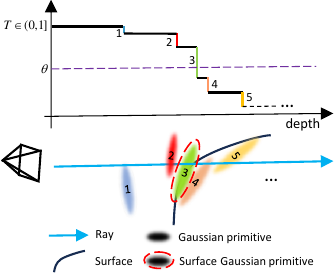}
\vspace{-3mm}
\caption{Surface Depth Analysis: The coordinate system illustrates the extent to which transmittance diminishes after a ray traverses the Gaussian, alongside a visualization of the Gaussian corresponding position. The parameter $\theta$ represents a predefined threshold; the Gaussian at which transmittance $T$ decreases to $\theta$ signifies the depth value nearest to the surface.}
\label{fig:surdepth}
\vspace{-6mm}
\end{figure}

\textbf{Surface depth.} 
Currently, the geometric feature tracking in 3DGS-SLAM~\cite{splatam,gsslam,sgsslam} is achieved through re-rendering depth, which has demonstrated promising results. However, the depth obtained through re-rendering is influenced by each Gaussian that the ray traverses, leading to potential instability. When a ray intersects a Gaussian with specific opacity, such as noise or artifacts (illustrated by index 1 in Fig.~\ref{fig:surdepth}), this contribution is also incorporated into the final rendering result. Inspired by the work of~\cite{2dgs,rtgslam,gspro}, we obtain the Gaussian that closely approximates the geometric surface by accumulated transmittance, which represents the surface depth:
\begin{equation}
{\tilde D_s} = \max \{ {z_i}|{T_i} > \theta \}.
\end{equation}
Since the surface depth is determined based on the median of the accumulated transmittance, if the transmittance remains below full transmittance 2$\theta$ ($\theta$ set to 0.5) even in a converged state, the surface depth will remain uncertain. Therefore, it is essential to estimate the accumulated transmittance of the pixel to ensure that the derived surface depth is stable.
\begin{equation}
\tilde{T} = \sum\limits_{i = 1}^n {2\theta\alpha _i}{T_i}.
\end{equation}

\subsection{Mapping}
\label{sub:Mapping}
\textbf{Initialization.} In SLAM, large-scale Gaussian primitives can occupy extensive spatial regions, potentially impeding the initialization of new Gaussians in areas that are under-reconstructed. To mitigate this issue, it is recommended to initialize the Gaussian scale to approximately the size of a single pixel~\cite{splatam,gaussianslam}. For the $i$-th Gaussian initialization, the position ${X^w_i}$ is determined by the 3D points ${X^c_i}$ in the camera coordinate system and the extrinsic matrix $W_{wc}$, which denotes the transformation from camera to world coordinates. The color ${c_i}$ is extracted from the RGB image, the opacity is initialized to 1, and the rotation matrix ${R_i}$ is initialized to the unit quaternion $q \in {SE(3)}$. The scale is initialized as the three-dimensional vector ${s_i} \in {\mathbb{R}^3}$ as follows:
\begin{equation}
{s_i} = \frac{{{{(K{{W_{wc}}}{X^c_i})}_z}}}{{{d_f}}},{\rm{   }}{d_f} = \left| {\frac{{{f_x} + {f_y}}}{2}} \right|,
\end{equation}
where $K$ is the camera intrinsic matrix, and ${(\cdot)_z}$ denotes the third dimension, corresponding to depth. Here, $f$ represents the camera focal length.

\begin{figure}[b]
    \centering
    \includegraphics[scale=0.55]{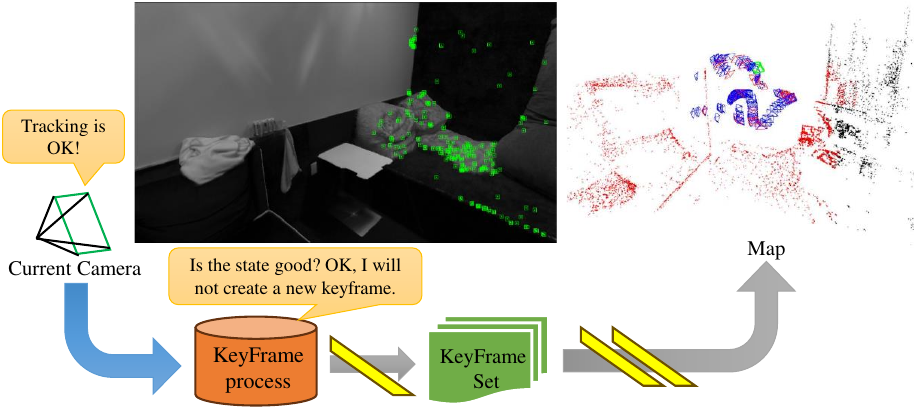}
    \vspace{-3mm}
    \caption{The green points represent feature points. At this moment, a keyframe should be generated, but in this case, since the tracking requirements are met, a keyframe is not generated and inserted into the map. As a result, observations of this region will be lost during subsequent training.}
    \label{fig:keyframe}
\end{figure}

\textbf{Adaptive densification.} To enhance the convergence speed of differentiable Gaussians and achieve high-fidelity results, it is essential to spawn new Gaussians that model the geometry and appearance of newly observed areas. We propose an appearance mask ${M_c}$, generated from the grayscale values of the re-rendered image, and a reconstructed geometric mask ${M_d}=(\left| {D - \tilde D} \right|<0.05)(\tilde D>0)$.

The geometric accuracy of rendering $\tilde D$ is contingent upon the convergence of differentiable Gaussians. Consequently, we establish an adaptive depth error threshold $\sigma_d$ to mitigate redundant additions:
\begin{equation}
{\sigma _d} = \mu \{ {E_d}\}  + 40 \eta \{ {E_d}\} 
\end{equation}
\begin{equation}
{E_d} = {\left| {D(x,y) - \tilde D(x,y)} \right|_{(x,y) \in {M_d}}}
\end{equation}
Here, $\mu\{\cdot\}$ denotes the mean operation, and $\eta\{\cdot\} $ signifies the median operation. New Gaussians will be spawned in regions that satisfy at least one of the following criteria:
\begin{equation}
Add({G_i}),\text{if }{(a)\text{ or }(b)}.
\end{equation}

Criterion $(a)$ is based on accumulated transmittance: ${\rm{\tilde T(x,y}}{\rm{)}}{\rm{ < 0}}{\rm{.8 }}$. This criterion facilitates improvements in areas where gradient descent is slow due to insufficient supervision by re-initializing them. Additionally, a new Gaussian must be generated if the accumulated transmittance has not reached 80\% of the maximum transmittance to ensure tracking accuracy.

Criterion $(b)$ is based on geometry: $({M_c}<50)({E_d} < {\sigma_d} )(\tilde T(x,y) < 0.99)$. The condition $({M_c}<50)({E_d} < {\sigma_d} )$ is employed to identify under-reconstructed areas by analyzing errors in geometry and appearance, while $(\tilde T(x,y) < 0.99)$ prevents the continuous addition of Gaussians due to significant geometric errors at the edges. As illustrated in Fig.~\ref{fig:pipeline}, the accumulated transmittance demonstrates greater robustness against edge errors.

\textbf{Keyframe generation.} 
Feature SLAM~\cite{orbslam2} primarily focuses on the quantity of feature points while neglecting their spatial distribution, as illustrated in Fig.~\ref{fig:keyframe}. The strategy may result in a lack of diverse training perspectives. To address this limitation, we propose an overlap graph method for generating keyframes that takes into account the degree of overlap, defined as follows:
\begin{equation}
{OG(i,j)} = \frac{{{\pi}({W_{ij},\mathcal{K}_j})}}{{{\mathcal{K}_i}}},
\end{equation}
where $\pi (\cdot)$ is projection function, $W_{ij}$ is the transformation matrix from frame $i$-th to frame $j$-th, and $\mathcal{K}_i$ indicates a set of 3D points in the $i$ frame. 

\textbf{Rendering frame selection mechanism.}
The rendering frames (RF) are selected from keyframes, and an effective selection mechanism can significantly enhance rendering quality. The TANBRIDGE~\cite{bridge} achieves remarkable reconstruction results by selecting a keyframe based on the co-visibility graph~\cite{orbslam2}. However, we have observed that relying exclusively on the set of rendering frames derived from the co-visibility graph may impose limitations on the training perspectives, potentially covering only a small portion of the primary object, as illustrated in Fig.~\ref{fig:keyframe}. This limitation may degrade the quality of reconstruction. To address this issue, we propose a novel rendering frame selection mechanism based on a hybrid graph.

Initially, we select the adjacency ${n_a}$ frames and the optimized ${n_b}$ frames to augment the RF set $\mathcal{R}$. The adjacency frames enhance the probability of training that includes the current viewpoint, as this viewpoint necessitates additional optimization for tracking. Furthermore, the optimized frames are added due to the presence of the backend pose optimization thread.

Secondly, to prevent the selection of nearly identical viewpoints into $\mathcal{R}$, particularly in scenarios of weak tracking, keyframes are continuously generated to improve tracking performance. Therefore, we initialize a window $W$ with the current frame $f_{1}$. 
Next, we employ a straightforward yet effective approach: If the candidate frame is derived from the co-visibility graph and meets condition $OG({f_{c}},{f_i}) < \beta_1$, it will be added to the set $\mathcal{R}$. This condition not only ensures the acquisition of additional viewpoints from common regions but also minimizes the inclusion of redundant viewpoints. Conversely, if the candidate frame is not belong to the co-visibility graph but satisfies condition $OG{({f_{1}},{f_{i}})}> \beta_2$, it will still be incorporated, as certain viewpoints may lack a co-visibility relationship based on feature points, yet still possess overlapping information with the current viewpoint. This procedure ensures diverse training viewpoints, thereby preventing overfitting. The pseudocode is provided in Alg.~\ref{alg:1}. 

\input{alg/RFcode}

Finally, ${n_r}$ frames are randomly selected and added to $\mathcal{R}$ to mitigate catastrophic forgetting of the global map due to continuous learning.
 
\textbf{Map optimization. } In frame-to-model tracking mode, map optimization is crucial. In continuous SLAM, we propose an isotropic regularization loss to overcome anisotropic influence (strip-shaped Gaussian primitives):
\begin{equation}
{{\cal L}_{iso}} =\frac{1}{N} \sum\limits_{i = 0}^N {(max\{ s_{i} \} -min \{ s_{i} \} |{s_i} > \gamma )},
\end{equation}
and a scalar regularization loss:
\begin{equation}
{{\cal L}_{s}} =\sum\limits_{i = 0}^N {( s_{i} - \gamma |{s_i} > \gamma )},\gamma = 0.03*\max \{ {({\bf{G}})_z}\}.
\end{equation}
To enhance the Gaussian fit to the surface, we introduce a surface depth loss:
\begin{equation}
{{\cal L}_{sur}} = {\left| {{{\tilde D}_s} - D} \right|_1}.
\end{equation}
Additionally, since the surface depth requires geometric depth for maintenance, we incorporate geometric supervision:
\begin{equation}
{{\cal L}_{d}} = {\left| {\tilde D - D} \right|_1}.
\end{equation}
For color supervision, we combine L1 and SSIM~\cite{ssim} losses:
\begin{equation}
{{\cal L}_{rgb}} = {\lambda }{\left| {\tilde C - C} \right|_1} + (1 - {\lambda })(1 - SSIM(\tilde C,C)).
\end{equation}
Final map optimization loss:
\begin{equation}
{{\cal L}_{mapping}} = {\omega _{1}^{m}}{{\cal L}_{rgb}} + {\omega _{2}^{m}}{{\cal L}_{d}} + {\omega _{3}^{m}}{{\cal L}_{sur}}+ {\omega _{4}^{m}}({{\cal L}_{iso}+2{\cal L}_{s}}),
\end{equation}
where $\omega^{m}$ is a set of weights for mapping.

\subsection{Tracking}
\label{sub:Tracking}

\textbf{Frame-to-model tracking.} We jointly optimize the photometric error derived from 3DGS re-rendering and ground truth, as well as the reprojection error based on feature points, to achieve accurate pose estimation. The reprojection loss is defined as follows:
\begin{equation}
{{\cal L}_{rpj}} = \sum\limits_{j\in \mathcal{M}} {\sum\limits_{i \in \mathcal{P}} {{\varphi}({{\Vert {{p_i} - \pi ({W_{\text{cj}}},{P^j_i})} \Vert}_2})} },
\end{equation}
where $\mathcal{M}$ denotes the set of local keyframes, $\mathcal{P}$ represents the matched features, $p_{i}$ is the pixel observation in the current image, $W_{\text{cj}}$ indicates
the pose of the $j$-th camera relative to the current camera and $P^{j}_{i}$ is the $i$-th 3D point observed by the $j$-th camera, and ${\varphi}$ represents the information matrix.

Building on the analysis presented in~\ref{sub:Preliminary}, we propose using surface depth instead of re-rendering depth to enhance tracking performance. Furthermore, we only utilize Gaussian surfaces with approximately complete transmittance as geometric features. Additionally, we optimize the combined loss using the Adam optimizer:
\begin{equation}
{{\cal L}_{tracking}} = (\tilde T > 0.99)({\omega _{1}^{t}}{{\cal L}_{rgb}} + {\omega _{2}^{t}}{{\cal L}_{sur}}) + {\omega _{3}^{t}}{{\cal L}_{rpj}}.
\end{equation}

During the optimization process, to prevent incorrect feature point matches from impacting the reduction of total loss, we will remove outliers in advance, allowing the re-rendering loss to be prioritized as the primary component.

\textbf{Bundle adjustment.} In the backend thread, we use graph optimization to jointly optimize map points and camera poses based on sparse feature points, similar to~\cite{orbslam2}. The backend operates as an independent thread, unaffected by the training. Therefore, once the camera poses are optimized, they can promptly provide more accurate camera poses for reconstruction, enhancing the reconstruction quality.

%% file: alg/RFcode.tex
\begin{algorithm}[t]\small
    \caption{Rendering frame selection mechanism}
    \label{alg:1}
    \renewcommand{\algorithmicrequire}{\textbf{Input:}}
    \renewcommand{\algorithmicensure}{\textbf{Output:}}
    \begin{algorithmic}[1]
        \REQUIRE the keyframe set $K$, the co-visibility graph of the current frame $CG$, the overlap graph $OG$, the co-visibility threshold $\beta_1$, the overlap threshold $\beta_2$, the current frame $f_1$, the maximum capacity $n_s$, the random number of frames $n_r$.
        \ENSURE rendering frame set $\mathcal{R}$
            \STATE Add adjacency $n_a$ frames and optimized $n_b$ frames  
            to $\mathcal{R}$.
            \STATE Initialize a window $W$.
            \STATE $f_{c}$ = $W$.front();
            \FOR{each $f_{i} \in K$}
                \IF {$f_{i} \in CG \And OG({f_{c}},{f_{i}} ) < \beta_1$} 
                \STATE $\mathcal{R}\text{.push}(f_{i})$;$W\text{.push}(f_{i})$;$f_{c}$ = $W\text{.front()}$;
                \ELSIF{$f_{i} \notin CG \And OG{({f_{1}},{f_{i}})} > \beta_2$}
                \STATE $\mathcal{R}\text{.push}(f_{i})$;$W\text{.push}(f_{i})$;
                \ENDIF
                \IF {$\mathcal{R}\text{.size()} > n_s$} 
                \STATE \text{break;}
                \ENDIF
            \ENDFOR
            \STATE Randomly select $n_r$ frames into $\mathcal{R}$;

        \RETURN $\mathcal{R}$
    \end{algorithmic}
\end{algorithm}

%% file: section/4_experiment.tex
\section{EXPERIMENT}
\subsection{Experiment Setup}
\textbf{Dataset.} To thoroughly evaluate the performance of the method presented in this paper, we select three sequences from the real-world TUM dataset~\cite{tum} and eight sequences from the synthetic Replica dataset~\cite{replica}, following the methodologies outlined in~\cite{splatam,niceslam,pointslam}.

\input{table/tum_ate1}

\textbf{Baesline.} We choose state-of-the-art NeRF SLAM methods, specifically NICE-SLAM~\cite{niceslam} and Point-SLAM~\cite{pointslam}. In addition, we include 3DGS-based SLAM methods such as SplatTAM~\cite{splatam}, GS-SLAM~\cite{gsslam}, MonoGS~\cite{gaussiansplattingslam}, and Sun~\cite{hf}. We also consider loosely coupled feature points with 3DGS methods, specifically Photo-SLAM~\cite{photoslam} and TAMBRIDGE~\cite{bridge}. Furthermore, we incorporate the traditional SLAM method ORB-SLAM2~\cite{orbslam2}.

\textbf{Metrics.} To measure RGB rendering performance, we utilize PSNR (dB), SSIM, and LPIPS. For camera pose estimation tracking, we employ the average absolute trajectory error (ATE RMSE [cm])~\cite{ate}. The \textbf{\color{red}{best}} results will be highlighted in red, while the \textbf{\color{blue}{second-best}} results will be highlighted in blue. The results represent the average of five trials.

\input{table/replica_ate}

\textbf{Implementation details.} We fully implement our SLAM algorithm in C++ and CUDA. Additionally, the SLAM algorithm runs on a desktop equipped with Intel i7-12700KF and an NVIDIA RTX 4060ti 16G GPU. We set parameters $\omega^{m}=\lbrace 1.0,0.7,0.1,5 \rbrace$,
$\lambda=0.8$,
$\lbrace \beta_1,\beta_2 \rbrace=\lbrace 0.07,0.3 \rbrace$,
$\lbrace n_a,n_b,n_s,n_r \rbrace=\lbrace 5,5,13,7 \rbrace$.
For the TUM datasets, we set weight $\omega^{t}=\lbrace 1.0,0.7,0.1 \rbrace$, 
and we set weight $\omega^{t}=\lbrace 0.7,1.0,0.1 \rbrace$ for the Replica datasets.

\input{table/depth_abl}

\subsection{Localization and Rendering Quality Evaluation}
\textbf{Localization.} Table~\ref{tab:atetum1} presents the tracking results on the TUM RGB-D dataset. Our method outperforms other baseline methods, including the state-of-the-art ORB-SLAM2. TANBRIDGE and Photo-SLAM are loosely coupled methods based on ORB-SLAM3~\cite{orbslam3}, while our tightly coupled method improves localization accuracy by an average of \textbf{26}\% compared to these methods. This improvement is attributed not only to the joint optimization of ORB feature and 3DGS but also to our novel surface geometry-based tracking method. Table~\ref{tab:atereplica} reports the results on the synthetic Replica dataset. The feature-based ORB-SLAM2 fails to track sequence R2 due to the presence of a weakly textured wall. In contrast, our approach not only achieves superior tracking across all sequences compared to ORB-SLAM2, resulting in a \textbf{16.2}\% improvement in RMSE, but also surpasses other 3DGS-based SLAM methods.

\input{table/replica_render}

\input{table/tum_render}

\input{table/ablation}
 \vspace{2mm}
\begin{figure*}[t]
		\centering
		\includegraphics[scale=0.6]{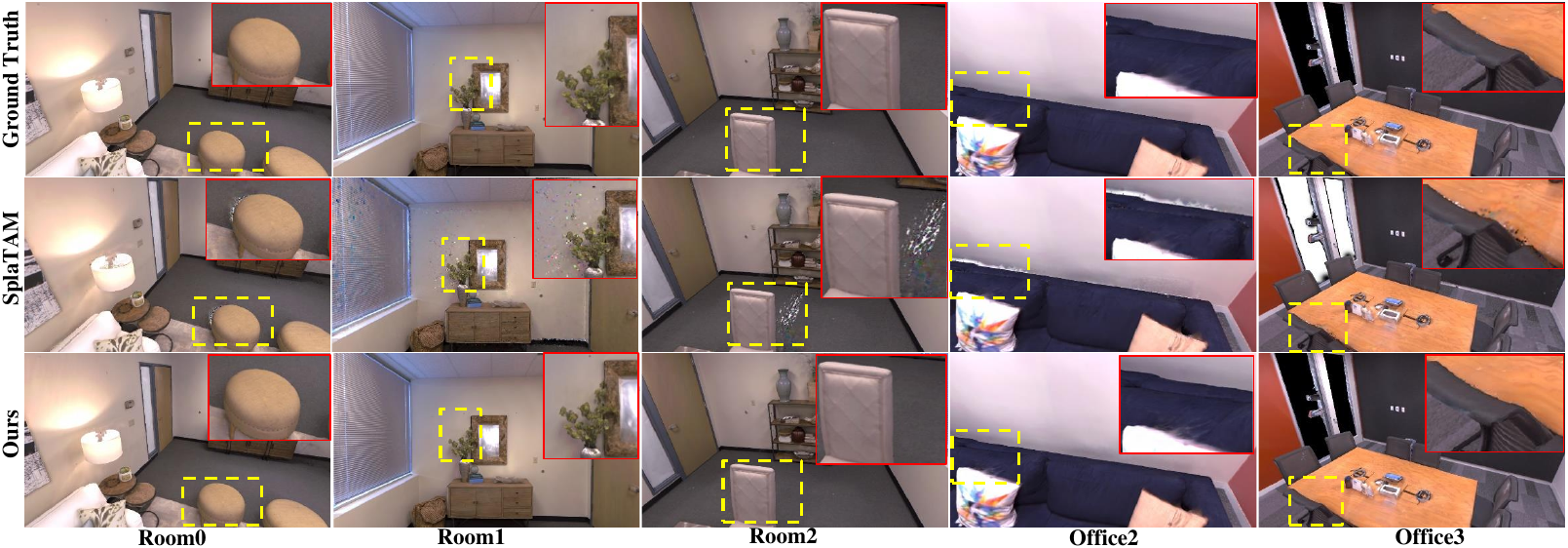}
  \vspace{-2mm}
\caption{The render visualization results on the Replica dataset.}
		\label{fig:vis}
  \vspace{-3mm}
	\end{figure*}

\textbf{Rendering quality.} 
We quantitatively evaluate rendering quality on the Replica and TUM RGB-D datasets, with results shown in Table~\ref{tab:renderreplica} and Table~\ref{tab:rendertum}, respectively. The results in Table~\ref{tab:renderreplica} indicate that our method improves PSNR by an average of \textbf{3.93} dB compared to other approaches. In Table~\ref{tab:rendertum}, our rendering quality improves by an average of \textbf{2.21} dB compared to loosely coupled methods, while also achieving a \textbf{1.16} dB improvement over 3DGS-based methods. This enhancement is attributed to the adaptive expansion and rendering frame selection mechanism we proposed. Visualization results in Fig.~\ref{fig:vis} demonstrate the ability of our method to generate higher-quality realistic images.

\subsection{Surface Depth Analysis}
We select the TUM dataset due to its inclusion of noise from real-world environments, effectively highlighting the advantages of our method. As shown in Table~\ref{tab:depth_ablation}, our surface depth method significantly outperforms the tracking method based on re-rendering depth. This is due to the ability of our method to reduce the impact of artifacts and noise on tracking, as illustrated in Fig.~\ref{fig:surdepth}. Furthermore, our method demonstrates superior speed compared to the re-rendering approach. This efficiency is achieved through our technique, which selects Gaussian primitives based on accumulated transmittance to approximate surface depth, thereby eliminating the need for re-rendering depth and enhancing the computation of geometric error gradients.

\subsection{Ablation Study}
To highlight the contributions of our components within GSORB-SLAM, we conduct a series of ablation experiments, with results averaged from the office sequence in Table~\ref{tab:ablation}. To make comparisons more meaningful, we refer to the co-visibility selection strategy. Additionally, to demonstrate the limitations of generating keyframes solely based on sparse feature points in constructing the 3D Gaussian representation, we compare the original keyframes generated by feature points in~\cite{orbslam2} with those produced by our method.

Through comprehensive analysis, we found that keyframe generation not only improves localization accuracy but also enhances rendering quality, achieving a \textbf{2} dB increase in PSNR. In contrast, the primary contribution of the other components lies in improving rendering quality. This improvement results from the combined effect of careful training viewpoint selection and regularization, which prevents the Gaussians from elongating into strip-like shapes.

%% file: table/tum_ate1.tex
\begin{table}[b]
    \vspace{-5mm}

     \renewcommand{\arraystretch}{1.2}
    \setlength{\tabcolsep}{2mm}
    \centering
    {\caption{Tracking Results on RGB-D TUM Dataset (ATE RMSE$\downarrow$ [cm]). "--" indicates unavailable data because the related work is not open. 
    \vspace{-3mm}
    \label{tab:atetum1}}}
    \begin{tabular}{c|cccc}
        \hline
        Method&Fr1/desk1&Fr2/xyz&Fr3/office&Avg. \\
        \hline
        ORB-SLAM2&1.60&0.40&\color{blue}\textbf{1.00}&\color{blue}\textbf{1.00}	\\
\hline
        Point-SLAM&4.34&1.31&3.48&3.04	\\

        NICE-SLAM&4.26&31.73&3.87&13.28	\\
 \hline
        MonoGS(RGB-D)&\color{blue}\textbf{1.52}&1.58&1.65&1.58	\\

        GS-SLAM&3.30&1.30&6.60&3.73	\\
  
        SplatTAM&3.35&1.24&5.16&3.25	\\

        Sun&3.38&--&5.12&--	\\
  
\hline
        Photo-SLAM&2.60&\color{blue}\textbf{0.35}&1.00&1.31	\\

        TANBRIDGE&1.75&\color{red}\textbf{0.32}&1.42&1.16	\\
\hline
        Ours&\color{red}\textbf{1.48}&{0.39}&\color{red}\textbf{0.88}&\color{red}\textbf{0.91}	\\
        \hline
    \end{tabular}
\end{table}

%% file: table/replica_ate.tex
\begin{table}[t]

\renewcommand{\arraystretch}{1.2}
    \setlength{\tabcolsep}{1.2mm}
    \centering
    {\caption{Tracking Results on the Replica Dataset (ATE RMSE$\downarrow$ [cm]). The Lost indicates tracking lost.
    \vspace{-3mm}
    \label{tab:atereplica}}}
    \begin{tabular}{c|cccccccc}
        \hline
        Method&R0&R1&R2&Of0&Of1&Of2&Of3&Of4 \\
        \hline
        ORB-SLAM2&0.45&\color{blue}\textbf{0.29}&\color{green}Lost&0.47&0.28&0.75&0.69&0.59	\\
        \hline
        Point-SLAM&0.61&0.41&0.37&\color{blue}\textbf{0.38}&0.48&0.54&0.69&0.72	\\

        NICE-SLAM&0.97&1.31&1.07&0.88&1.00&1.06&1.10&1.13	\\
        \hline
        MonoGS(RGB-D)&0.47&0.43&0.31&0.70&0.57&\color{blue}\textbf{0.31}&\color{red}\textbf{0.31}&3.20	\\

        GS-SLAM&0.48&0.53&0.33&0.52&0.41&0.59&0.46&0.70	\\
        SplatTAM&\color{red}\textbf{0.31}&0.40&\color{red}\textbf{0.29}&0.47&\color{blue}\textbf{0.27}&\color{red}\textbf{0.29}&\color{blue}\textbf{0.32}&\color{blue}\textbf{0.55}	\\
        \hline
        Ours&\color{blue}\textbf{0.35}&\color{red}\textbf{0.22}&\color{blue}\textbf{0.33}&\color{red}\textbf{0.33}&\color{red}\textbf{0.19}&0.54&0.63&\color{red}\textbf{0.45}	\\
        \hline
    \end{tabular}
		    \vspace{-6mm}
\end{table}

%% file: table/depth_abl.tex
\begin{table}[b]
    \vspace{-5mm}

\renewcommand{\arraystretch}{1}
    \setlength{\tabcolsep}{1mm}
	\centering
 	\caption{Tracking results at different depths on the TUM datasets. (ATE RMSE$\downarrow$ [cm])}
        \vspace{-3mm}
	\label{tab:depth_ablation}
	\begin{tabular}{c|cccc}
		\hline
		Method&Fr1/desk1&Fr2/xyz&Fr3/office&Tracking/Iter.(ms)  \\
  \hline
Re-rendering depth &2.46&0.41&0.94 &16        \\
Surface depth (Ours) &\color{red}\textbf{1.48}&\color{red}\textbf{0.39}&\color{red}\textbf{0.88} &\color{red}\textbf{11}          \\ 
  \hline
	\end{tabular}
\end{table}

%% file: table/replica_render.tex
\begin{table}[t]
   \vspace{-1mm}
\renewcommand{\arraystretch}{1.3}
	\centering
         {\caption{Rendering performance comparison of RGB-D SLAM methods on Replica.}\label{tab:renderreplica}}
         \vspace{-3mm}
         \setlength{\tabcolsep}{0.3mm}
	\begin{tabular}{c|ccccccccc}
		\hline
Sequence & Metric & R0    & R1    & R2    & Of0   & Of1   & Of2   & Of3   & Of4   \\ \hline
\multirow{3}{*}{Point-SLAM}
& PSNR$\uparrow$   & 32.40 & 34.80 & \color{blue}\textbf{35.50} & 38.26 & 39.16 & \color{blue}\textbf{33.99} & \color{blue}\textbf{33.48} & \color{blue}\textbf{33.49}       \\
& SSIM$\uparrow$   & 0.97  & \color{blue}\textbf{0.98}  & 0.98  & 0.98  & 0.99  & 0.96  & 0.96  & \color{blue}\textbf{0.98}        \\
& LPIPS$\downarrow$  & 0.11  & 0.12  & 0.11  & 0.10  & 0.12  & 0.16  & 0.13  & 0.14        \\ \hline
\multirow{3}{*}{NICE-SLAM} 
& PSNR$\uparrow$   & 22.12 & 22.47 & 24.52 & 29.07 & 30.34 & 19.66 & 22.23 & 24.49       \\
& SSIM$\uparrow$   & 0.69  & 0.76  & 0.81  & 0.87  & 0.89  & 0.80  & 0.80  & 0.86        \\
& LPIPS$\downarrow$  & 0.33  & 0.27  & 0.21  & 0.23  & 0.18  & 0.24  & 0.21  & 0.20        \\ \hline
\multirow{3}{*}{SplatTAM}      
& PSNR$\uparrow$   & \color{blue}\textbf{32.86} & \color{blue}\textbf{33.89} & 35.25 & 38.26 & 39.17 & 31.97 & 29.70 & 31.81       \\
& SSIM$\uparrow$   & \color{blue}\textbf{0.98}  & 0.97  & \color{blue}\textbf{0.98}  & 0.98  & 0.98  & 0.97  & 0.95  & 0.95        \\
& LPIPS$\downarrow$  & \color{blue}\textbf{0.07}  & 0.10  & \color{blue}\textbf{0.08}  & 0.09  & 0.09  & 0.10  & 0.12  & 0.15        \\ \hline
\multirow{3}{*}{GS-SLAM}      
& PSNR$\uparrow$   & 31.56 & 32.86 & 32.59 & \color{blue}\textbf{38.70} & \color{blue}\textbf{41.17} & 32.36 & 32.03 & 32.92       \\
& SSIM$\uparrow$   & 0.968 & 0.973 & 0.971 & \color{blue}\textbf{0.986} & \color{blue}\textbf{0.993} & \color{blue}\textbf{0.978} & \color{blue}\textbf{0.970} & 0.968       \\
& LPIPS$\downarrow$  & 0.094 & \color{blue}\textbf{0.075} & 0.093 & \color{blue}\textbf{0.050} & \color{blue}\textbf{0.033} & \color{blue}\textbf{0.094} & \color{blue}\textbf{0.110} & \color{blue}\textbf{0.112}       \\ \hline

\multirow{3}{*}{Ours}          
& PSNR$\uparrow$   &\color{red}\textbf{35.46} &\color{red}\textbf{37.96} &\color{red}\textbf{38.43} &\color{red}\textbf{41.89} &\color{red}\textbf{42.05}&\color{red}\textbf{36.27}  &\color{red}\textbf{35.86} &\color{red}\textbf{37.04}      \\
& SSIM$\uparrow$   &\color{red}\textbf{0.986} &\color{red}\textbf{0.990} &\color{red}\textbf{0.992} &\color{red}\textbf{0.993}  &\color{red}\textbf{0.993} &\color{red}\textbf{0.988} &\color{red}\textbf{0.989}    &\color{red}\textbf{0.989}       \\
& LPIPS$\downarrow$  &\color{red}\textbf{0.037} &\color{red}\textbf{0.039} &\color{red}\textbf{0.038}&\color{red}\textbf{0.034} &\color{red}\textbf{0.030} &\color{red}\textbf{0.050} &\color{red}\textbf{0.041}     &\color{red}\textbf{0.053}       \\ \hline
	\end{tabular}
 \vspace{-6mm}
\end{table}

%% file: table/tum_render.tex
\begin{table}[t]
   \vspace{-1mm}
\renewcommand{\arraystretch}{1}
	\centering
 {\caption{Rendering performance comparison of RGB-D SLAM methods on TUM Datasets. "--" indicates unavailable data because the related work is not open. \label{tab:rendertum}}}
 \vspace{-3mm}
         \setlength{\tabcolsep}{3mm}
	\begin{tabular}{c|cccc}
		\hline
Sequence & Metric & Fr1/desk1 & Fr2/xyz & Fr3/office \\ \hline
\multirow{3}{*}{Point-SLAM} 
& PSNR$\uparrow$   & 13.87     & 17.56   & 18.43      \\
& SSIM$\uparrow$   & 0.63      & 0.71    & 0.75       \\
& LPIPS$\downarrow$  & 0.54      & 0.59    & 0.45       \\ \hline
\multirow{3}{*}{NICE-SLAM}  
& PSNR$\uparrow$   & 12.00     & 18.20   & 16.34      \\
& SSIM$\uparrow$   & 0.42      & 0.60    & 0.55       \\
& LPIPS$\downarrow$  & 0.51      & 0.31    & 0.39       \\ \hline
\multirow{3}{*}{TANBRIDGE}  
& PSNR$\uparrow$   & 21.22     & 23.44   & 20.15      \\
& SSIM$\uparrow$   & 0.88      & 0.90    & 0.82       \\
& LPIPS$\downarrow$  & 0.19      & \color{red}\textbf{0.10}    & 0.25       \\ \hline
\multirow{3}{*}{Photo-SLAM} 
& PSNR$\uparrow$   & 20.870    & 22.094  & \color{blue}\textbf{22.744}     \\
& SSIM$\uparrow$   & 0.743     & 0.765   & 0.780      \\
& LPIPS$\downarrow$  & 0.239     & 0.169   & \color{red}\textbf{0.154}      \\ \hline
\multirow{3}{*}{Sun}        
& PSNR$\uparrow$   & \color{blue}\textbf{22.60}     & --      & 22.30      \\
& SSIM$\uparrow$   & \color{red}\textbf{0.91}      & --      & \color{blue}\textbf{0.89}       \\
& LPIPS$\downarrow$  & \color{red}\textbf{0.15}      & --      & \color{blue}\textbf{0.16}       \\ \hline
\multirow{3}{*}{SplatTAM}   
& PSNR$\uparrow$   & 22.00      & \color{blue}\textbf{24.50}   & 21.90      \\
& SSIM$\uparrow$   & 0.86      & \color{red}\textbf{0.95}    & 0.88       \\
& LPIPS$\downarrow$  & 0.19      & 0.10    & 0.20       \\ \hline
\multirow{3}{*}{Ours}       
& PSNR$\uparrow$   &\color{red}\textbf{23.02}   &\color{red}\textbf{24.78}    &\color{red}\color{red}\textbf{24.08}            \\
& SSIM$\uparrow$   &\color{blue}\textbf{0.887}    &\color{blue}\textbf{0.935}   &\color{red}\textbf{0.914}            \\
& LPIPS$\downarrow$  &\color{blue}\textbf{0.176}  &\color{blue}\textbf{0.114}   &0.171            \\ \hline
	\end{tabular}
 \vspace{-3mm}
\end{table}

%% file: table/ablation.tex
\begin{table}[t]
    \renewcommand{\arraystretch}{1}
    \setlength{\tabcolsep}{2mm}
	\centering
  	\caption{The Ablation Analysis on Replica Office sequences. The result is the average value of all office sequences.}
        \vspace{-3mm}
	\label{tab:ablation}
		\begin{tabular}{ccc|cc}
  \hline
		\multicolumn{3}{c|}{Variable}         & \multicolumn{2}{c}{Average}         \\
		RF Mech. & KF Gen.    & Reg.       & ATE $\downarrow$    & PSNR $\uparrow$        \\ \hline
		\Checkmark           & \Checkmark          & \Checkmark  
        & \color{red}\textbf{0.422}        & \color{red}\textbf{39.01}               \\
		\XSolidBrush           & \Checkmark          & \Checkmark
        & 0.425        & 37.32              \\
		\Checkmark           & \XSolidBrush          & \Checkmark 
        & 0.551        & 36.95               \\
		\Checkmark           & \Checkmark          & \XSolidBrush 
        & 0.430        & 38.51            \\ \hline  
	\end{tabular}
       
\vspace{-6mm}
\end{table}

%% file: section/5_conclusion.tex
\section{CONCLUSION}
In this paper, we introduced GSORB-SLAM, a tightly coupled system that integrates 3DGS with ORB features. Experimental results demonstrated that our joint optimization method, which leverages geometric surfaces and feature points, effectively reduces the system sensitivity to noise. Additionally, we developed a viewpoint selection strategy based on a hybrid graph and an adaptive Gaussian expansion method to enhance the rendering quality of dense SLAM. Our experiments showcased the impressive performance of the proposed approach. However, the method currently necessitates a considerable amount of time for training and is unable to perform real-time localization, highlighting areas for future research.